\documentclass{article}

\usepackage{arxiv}

\usepackage[utf8]{inputenc} 
\usepackage[T1]{fontenc}    
\usepackage{hyperref}       
\usepackage{url}            
\usepackage{booktabs}       
\usepackage{amsfonts}       
\usepackage{nicefrac}       
\usepackage{microtype}      
\usepackage{lipsum}
\usepackage{graphicx}
\graphicspath{ {./images/} }

\usepackage[numbers]{natbib}

\usepackage[english]{babel}

\usepackage{amsmath,amssymb,amsfonts}
\usepackage{algorithmic}
\usepackage{graphicx}
\usepackage{textcomp}
\usepackage{xcolor}
\usepackage{selinput}
\usepackage{parskip}
\usepackage{booktabs}
\usepackage{hyperref}
\usepackage{multicol}
\usepackage{multirow}
\usepackage{makecell}
\usepackage{csquotes}
\usepackage{adjustbox}

\usepackage{mathtools}
\usepackage{amsmath}
\usepackage{verbatim}
\usepackage{subcaption}
\usepackage{graphicx}
\usepackage{caption}

\title{An Efficient Gradient-Based Inference Attack for Federated Learning}

\author{
 Pablo Montaña-Fernández \\
  Gradiant\\
  Carretera do Vilar 56-58, 36214, Vigo, Spain \\
  \texttt{pmontana@gradiant.org} \\
   \And
 Ines Ortega-Fernandez \\
  Gradiant\\
  Carretera do Vilar 56-58, 36214, Vigo, Spain \\
  \texttt{iortega@gradiant.org}
}

\begin{document}
\maketitle
\begin{abstract}
Federated Learning is a machine learning setting that reduces direct data exposure, improving the privacy guarantees of machine learning models. Yet, the exchange of model updates between the participants and the aggregator can still leak sensitive information. In this work, we present a new gradient-based membership inference attack for federated learning scenarios that exploits the temporal evolution of last-layer gradients across multiple federated rounds. Our method uses the shadow technique to learn round-wise gradient patterns of the training records, requiring no access to the private dataset, and is designed to consider both semi-honest and malicious adversaries (aggregators or data owners). Beyond membership inference, we also provide a natural extension of the proposed attack to discrete attribute inference by contrasting gradient responses under alternative attribute hypotheses. The proposed attacks are model-agnostic, and therefore applicable to any gradient-based model and can be applied to both classification and regression settings. We evaluate the attack on CIFAR-100 and Purchase100 datasets for membership inference and on Breast Cancer Wisconsin for attribute inference. Our findings reveal strong attack performance and comparable computational and memory overhead in membership inference when compared to another attack from the literature. The obtained results emphasize that multi-round federated learning can increase the vulnerability to inference attacks, that aggregators pose a more substantial threat than data owners, and that attack performance is strongly influenced by the nature of the training dataset, with richer, high-dimensional data leading to stronger leakage than simpler tabular data.
\end{abstract}


\section{Introduction}

In Spain, it is estimated that approximately 286,664 new cancer cases will be diagnosed in 2024 \cite{cifras-cancer-2024}. This implies that, when selecting a random person from the total population and assuming independence from risk factors (such as age, smoking habits, medical history, etc.), the probability that they have been diagnosed with cancer during that year is approximately $0.6\%$, or 1 in 166 people.

Consider a statistical model trained on clinical data from individuals diagnosed with cancer, along with data from healthy individuals (to learn to distinguish between them). These models may be vulnerable to privacy attacks such as membership inference attacks (MIAs), which aim to determine whether a specific record was included in the training dataset. While at first glance determining membership might seem unimportant, its relevance can be illustrated with the following example: suppose an adversary with access to the model parameters carries out an MIA. If the attack succeeds and the adversary can determine with high confidence that a particular sample was in the training data, it is indirectly revealed that the person associated with that sample was included in the medical dataset. Now, if some dataset details are publicly available and the proportion of cancer patients is known to exceed $0.6\%$—which is reasonable given that classes in training data are typically relatively balanced—the prior probability that the identified training member has cancer would increase. This substantial probability change highlights the real privacy risks of training machine learning models on sensitive data without applying proper protection mechanisms.

Privacy risks are generating a major debate about information security and privacy, as well as the use of personal data for training artificial intelligence models. In this regard, recent years have seen advances in the regulation of personal data use, with frameworks such as the General Data Protection Regulation (GDPR) \cite{GDPR}, aimed at protecting the processing of personal data of individuals, and the AI Act \cite{ai_act}, which seeks to regulate the legitimate uses of artificial intelligence and machine learning.

As a result, the current trend is to enhance the privacy guarantees offered by artificial intelligence algorithms. In this context, the \textit{Federated Learning (FL)} methodology \cite{google-federated-learning} emerged—a training schema in which the goal is to jointly train a model across different users or entities without the need to share training data in a centralized data repository. Each user or entity trains their own model locally and then shares it with a central server or aggregator, where a global model is built incorporating all participants' contributions. This process is repeated over several rounds until the global model converges. In this way, the global model benefits from the knowledge and learning of each local model, decentralizing training and avoiding the sharing of sensitive data between participants.

Although federated learning improves the privacy guarantees of machine learning models, these models remain vulnerable to different privacy attacks—such as MIAs— that aim to exploit information embedded in the trained model and its parameters to extract details about the training dataset, introduce perturbations, or even recover the original training data. While such attacks have been studied in centralized model training, they pose a greater privacy risk in federated settings \cite{nasr2019comprehensive} due to several distinctive characteristics. First, each participant has white-box access to the shared models, enabling more extensive computations and analyses. Second, participants receive different versions of these models after updates with private data, allowing them to study the evolution of the models across rounds. Finally—and most critically—each client actively participates in the training process, which creates opportunities for malicious behavior aimed at extracting additional information about private datasets through model updates.

In this work, we introduce a novel gradient-based membership inference attack for federated learning environments. The approach leverages membership-related information embedded in the model’s last-layer gradients to train a binary logistic regression classifier that distinguishes between training and non-training data. The proposed attack uses the shadow-training technique to exploit the temporal evolution of gradients across multiple FL rounds and requires no access to the private training records themselves, allowing for both semi-honest and malicious adversaries (aggregators or data owners). Beyond membership inference, we show how the attack can be naturally adapted to perform attribute inference. The proposed attack achieves high performance while introducing two key contributions:
\begin{itemize}
    \item \textbf{Computational efficiency.} This approach relies exclusively on the norm of the last-layer gradients, condensing gradient information into a single scalar value. This significantly reduces the size of the datasets used for training an attack model and enables the use of lightweight classifiers. In fact, we employ logistic regression, which is far more efficient than the complex models used in related work (e.g., [6]) for both training and inference, while maintaining high effectiveness. This fact makes the method practical for large-scale, multi-round FL scenarios.

    \item \textbf{Model-agnostic applicability.} By focusing exclusively on gradients, our approach is applicable to any training scenario using gradient-based optimization (including regression and gradient-boosting models), whereas other methods are limited to classification tasks because they rely on class-probability vectors.
\end{itemize}
To empirically evaluate the performance of the attack, we evaluate our method and compare it to another common MIA from the literature on the CIFAR-100, Purchase100, and Breast Cancer Wisconsin datasets, and study both the success rate of the attack and the computational efficiency.

The remaining of this paper is structured as follows: Section~\ref{sec:sota} surveys membership and attribute inference attacks in both centralized and federated settings. Section~\ref{sec:methods} formalizes our threat models and methodology, introducing MIA fundamentals (Sec.~\ref{sec:membershipinference}), the shadow training technique (Sec.~\ref{sec:shadowtechnique}), and its adaptation to federated learning (Sec.~\ref{sec:shadowFL}), and outlines the gradient-based and attribute inference ideas that motivate the design of our proposal. Section~\ref{sec:evaluation} details the experimental setup, including the datasets used to evaluate the attack performance, the FL settings, adversary types, and implementation choices. Section~\ref{sec:results} reports the quantitative results achieved by the proposed gradient-based attack, comparing it to another common MIA from the literature, and introduces the attribute inference evaluation. Finally, Section~\ref{sec:conclusions} summarizes the main conclusions and highlights directions for future work.

\section{Related Work}\label{sec:sota}

In this section, we examine the most relevant practical membership and attribute inference attacks for both centralized and federated learning scenarios. 

Based on the knowledge the attacker has about the target model, attacks can be widely classified into two main types: white-box and black-box attacks. In white-box attacks, the attacker has complete information, both from the training dataset and the target model, enabling them to know the values obtained across the different layers of the model, as well as the resulting output. In contrast, in a black-box scenario, the attacker only interacts with the target model as if it were a black box: they send a prediction request and receive the outcome produced by the model (in the form of a regression value, the results of a binary classification, etc.). In this case, the attacker may or may not have prior information about the training dataset. Black-box attacks are conducted with less information than white-box attacks, but if successful, they are far more dangerous due to presenting a significant privacy breach, as the attacker can compromise the model’s privacy with limited access to it.

To provide a comprehensive understanding and context of the previous work, this section surveys the two types of attacks implemented: membership inference and attribute inference attacks, that aim to determine whether a specific data point was part of the training set or to deduce sensitive attributes.

The membership inference attack has received considerable attention, with numerous works formalizing it and quantifying inference risk \cite{dwork2006differential, yeom2018privacy}, proposing practical attacks \cite{SSSS17, JWKGE20, zari2021efficient}, or designing mitigations \cite{abowd20222020}. In realistic scenarios, the adversary does not have access to the exact training data. Instead, the attack relies on obtaining samples that are statistically representative or similar to the target dataset. The effectiveness of the attack depends on multiple factors, including the model architecture, training methods, and the distribution of the underlying training data.

Authors in \cite{nasr2019comprehensive} propose both passive and active white-box membership inference attacks by exploiting access to model parameters, gradients and outputs; the attacker uses internal representations and training dynamics to distinguish members from non-members in centralized and federated learning settings. Their results show that federated learning can mitigate some privacy risks compared to centralized learning, yet FL remains vulnerable to membership inference attacks. The work also proposes several defense mechanisms, including differential privacy and adversarial training, to improve the privacy of deep learning models. Although this attack achieves strong results in the inference of membership, it is computationally expensive. For each record, the adversary must compute the outputs of all layers of the target model, the gradients with respect to each parameter, and the loss function value. Since training the attack model requires a large number of records, this makes the method infeasible in large-scale scenarios. \cite{Truex2019} further analyzed these attacks in the context of Machine Learning as a Service (MLaaS), showing that attackers can query models and use confidence scores or prediction entropy to infer membership, even without internal access, by training shadow models to mimic the target. Later, \cite{carlini2022membership} proposed an attack which improved evaluation by focusing on recall at low false-positive rates, revealing that many prior attacks were less effective than initially believed. The paper formalizes the attack as a likelihood ratio test using model predictions and loss values. Collectively, these studies underscore the significant privacy risks that membership inference poses in traditional machine learning environments.

Compared to centralized settings, the study of membership inference attacks in federated learning remains relatively limited. 
In contrast to prior work \cite{nasr2019comprehensive} which were computationally expensive,
\cite{zari2021efficient} proposed a new attack in the FL setting that is both less computationally demanding and more memory-efficient, while maintaining good accuracy. Their approach only feeds the attack model with the output probability of the ground-truth class for each record. However, in many black-box settings this probability is unavailable to the adversary. Moreover, they assume access to part of the training dataset, which is often unrealistic, making this approach less practical. Later, \cite{Gu2022} proposed \textit{Confidence Series based Membership Inference Attack} (CS-MIA), an attack based on prediction confidence series and shadow models, aiming to overcome previous limitations by relaxing the assumption of access to part of the training dataset, and considering an adversary with access to the population distribution from which the training data was sampled.

Finally, in attribute inference attacks, the adversary aims to learn unknown features of a user from the training set. For example, partial knowledge of a user’s responses to a questionnaire may allow the adversary to infer answers to other sensitive questions by querying a model trained on these responses. Given access to incomplete input records (which may be public), the adversary attempts to infer the missing private attributes. Attribute inference attacks in federated learning (FL) can be categorized into two groups: gradient-based or model-based attacks. 

In gradient-based attacks \cite{LC21} the adversary leverages the gradients shared during the FL training process to infer sensitive attributes by minimizing the distance between the real gradients and virtual ones. The effectiveness of this approach depends on the dataset size of the data owners (DOs): since attribute inference targets individual samples, larger datasets reduce the influence of a single record on the aggregated gradients.

On the other side, model-based attacks \cite{DXNGT22} where the adversary directly exploits the models exchanged during the FL rounds in order to reconstruct the local target model that one of the DOs trained on private data, therefore revealing sensitive attributes of the underlying training dataset.

Despite these recent advances, existing inference attacks still face constraints in terms of generality and applicability. In contrast, our work introduces an attack that not only mitigates these limitations but also offers greater versatility than CS-MIA, as it can be applied to any optimization problem involving gradient-based methods, and can be naturally extended to an attribute inference attack.

\section{Methodology}\label{sec:methods}

In this section, we formalize the methodological background of our work by presenting membership inference attacks and the shadow training technique, which together establish the basis for the proposed federated learning attack.

\subsection{Membership Inference Attacks}
\label{sec:membershipinference}

As previously introduced, a membership inference attack allows an adversary to infer whether a particular data record was used to train a specific machine learning (ML) model. For example, as previously discussed, participation in a particular clinical trial may reveal sensitive private information about a diagnosis.

An external attacker can conduct such attacks even without access to the model parameters (black-box scenario), simply by querying the model during inference and observing its output, obtaining the prediction vector $f(x) = (p_1(x),\ldots, p_k(x))$ for an arbitrary input $x$, where $p_i(x)$ is the probability of $x$ belonging to class $i$. These black-box attacks exploit a common trait of well-trained models: they tend to perform better on training data than on unseen examples. The adversary does not need to know the internal parameters of the target model—access to the trained model is sufficient to apply one or more techniques to indirectly assess if a sample belongs to the training set. Two widely used options to measure prediction confidence are:

\begin{itemize}
    \item True Class Prediction Probability: if $x$ belongs to class $y$, use $p_y(x)$. Since the model is trained to maximize $p_y(x)$ for training inputs of class $y$, this value is typically higher for samples that the model has already seen during training.
    \item Prediction Entropy: using the full prediction output, one can compute $-\sum_{i=1}^k p_i(\mathbf{x})\text{log }p_i(\mathbf{x})$. To express confidence (and increase values for members), it is common to use the negative entropy $\sum_{i=1}^k p_i(\mathbf{x})\text{log }p_i(\mathbf{x})$.
\end{itemize}

\begin{figure}[!htb]
    \centering \includegraphics[width=0.9\linewidth]{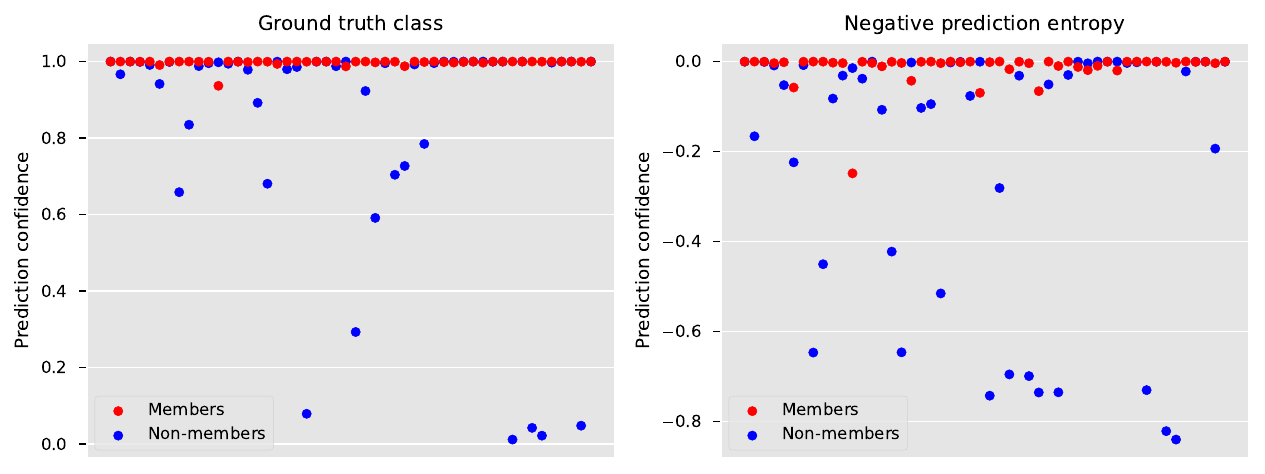}
    \caption{Prediction confidence differences between 50 random members and 50 non-members using true class prediction probability and negative prediction entropy on the Purchase100 dataset.}
    \label{fig:staticconfidence}
\end{figure}

Fig. \ref{fig:staticconfidence} illustrates the differences in these prediction confidence metrics when computed for training samples compared to non-training samples.

In a federated learning setting, and assuming an internal adversary, the attacker has white-box access to the complete model. The attacker could be the central aggregator, observing individual updates from each \textit{Data Owner} (DO) over time, or one of the DOs, observing aggregated global updates. In both cases, the adversary has access to the model's architecture and all its parameters, allowing the use of all this information when executing the attack. In this context, we can distinguish three types of behaviors from each participant in the federation:

\begin{itemize}
    \item \emph{Honest.} An honest participant strictly follows the federated learning protocol without any deviation. They do not attempt to infer private information or manipulate the training process. This behavior represents the baseline assumption in most FL systems.

    \item \emph{Semi-Honest.} A semi-honest adversary follows the prescribed protocol but attempts to extract additional information from the data they legitimately receive. For example, they may analyze gradients or model updates to infer sensitive information about other participants' data, even though they do not disrupt the training process.

    \item \emph{Malicious.} A malicious adversary actively deviates from the protocol to achieve their objectives. This may involve injecting poisoned updates, manipulating gradients, or colluding with other parties to compromise the integrity of the model or extract private data. Malicious behavior poses the highest risk to privacy.
\end{itemize}

In addition to the central aggregator and individual DOs, other potential adversaries in federated learning include external attackers who intercept communication channels and colluding participants who share local updates to enhance inference capabilities. External attackers, however, generally lack access to model parameters or training dynamics, which are essential inputs for our proposed approach. Colluding participants could theoretically combine updates to improve attack success, but this scenario lies outside our scope. Our work focuses on non-colluding entities extracting membership information directly from the models they access, which represents a realistic and widely studied threat model.

\subsection{Shadow Technique}
\label{sec:shadowtechnique}

As mentioned, differences in the target model's behavior between members and non-members can be leveraged for membership inference attacks. However, if the adversary lacks access to the training set, analyzing such differences is not possible. To address this, \cite{SSSS17} proposed the ``shadow training'' technique, which has shown great success in membership inference attacks.

The core idea is to replicate the training process of the target model using a known dataset and a ``shadow model'' with the same architecture. Since we know which records are used for training, we can use the shadow model to observe behavioral differences between members and non-members. Although \cite{SSSS17} also proposes a specific membership attack, the shadow training technique is adaptable to various scenarios. Several researchers have proposed modifications and variants (e.g., \cite{Salem2018}, \cite{Long2020}, \cite{Truex2019}, \cite{Song2019}), giving rise to an entire family of such attacks.

The general approach is as follows: suppose we want to attack a model $M$ trained on a private dataset $D_{\text{target}}$. If we know the distribution of $D_{\text{target}}$, we can obtain a ``shadow set'' $D_m$ (members) to simulate it (or, if we have partial access to $D_{\text{target}}$, we can build $D_m$ from this and additional data from the same distribution).

Then, we construct a shadow model $S$ with the same architecture as $M$ and train it on $D_m$. Once trained, we use a disjoint dataset $D_{nm}$ (non-members) to analyze the model’s behavior on unseen data. Since $S$ and $D_m$ mimic $M$ and $D_{\text{target}}$, the behavior difference between $D_m$ and $D_{nm}$ should mirror the difference between members and non-members of $D_{\text{target}}$.

To find such differences, the shadow technique analyzes computations performed by the target model based on whether an input was used during training. For example, the MIA proposed in \cite{nasr2019comprehensive} uses gradients, activations, loss values, and labels from the neural network to train a model that distinguishes members from non-members. This is one of the most widespread and effective attacks, though it is computationally expensive. On the other hand, works like \cite{zari2021efficient, Truex2019} show that similar or better results can be achieved using only the model output as the distinguishing factor.

\subsection{Shadow Tecnhique in Federated Learning}
\label{sec:shadowFL}

\begin{figure}[!htb]
\centering \includegraphics[width=0.9\linewidth]{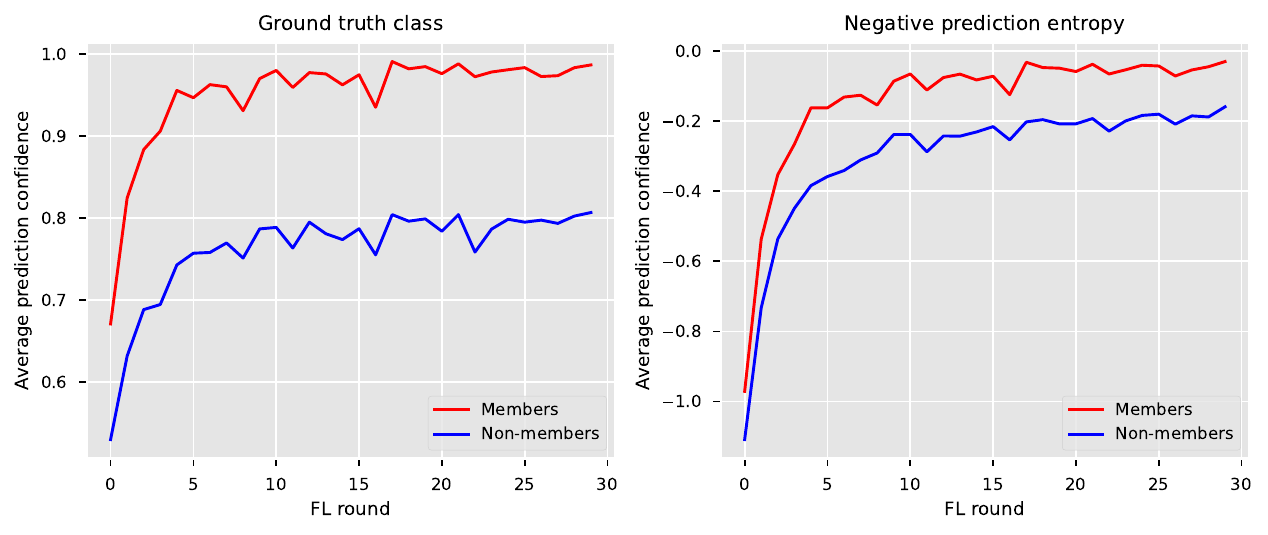}
\caption{Average prediction confidence of training set members and non-members from a specific data owner in FL over 30 rounds using both presented confidence measures. Dataset: Purchase100.}
\label{fig:flconfidence}
\end{figure}

In federated learning environments, participants can not only access a single model but also observe its temporal evolution over multiple rounds. This time-based behavior also differs between members and non-members (Fig.~\ref{fig:flconfidence}), making MIAs particularly suitable in federated learning scenarios.

To clarify the shadow technique scheme in this case, consider a scenario with $n$ federated training rounds, producing target models $M_1, \ldots, M_n$ updated each round using the private local dataset $D_{\text{target}}$ of the target data owner.

\paragraph{Training Phase}

\begin{figure*}[!t]
  \centering
  \includegraphics[width=\textwidth]{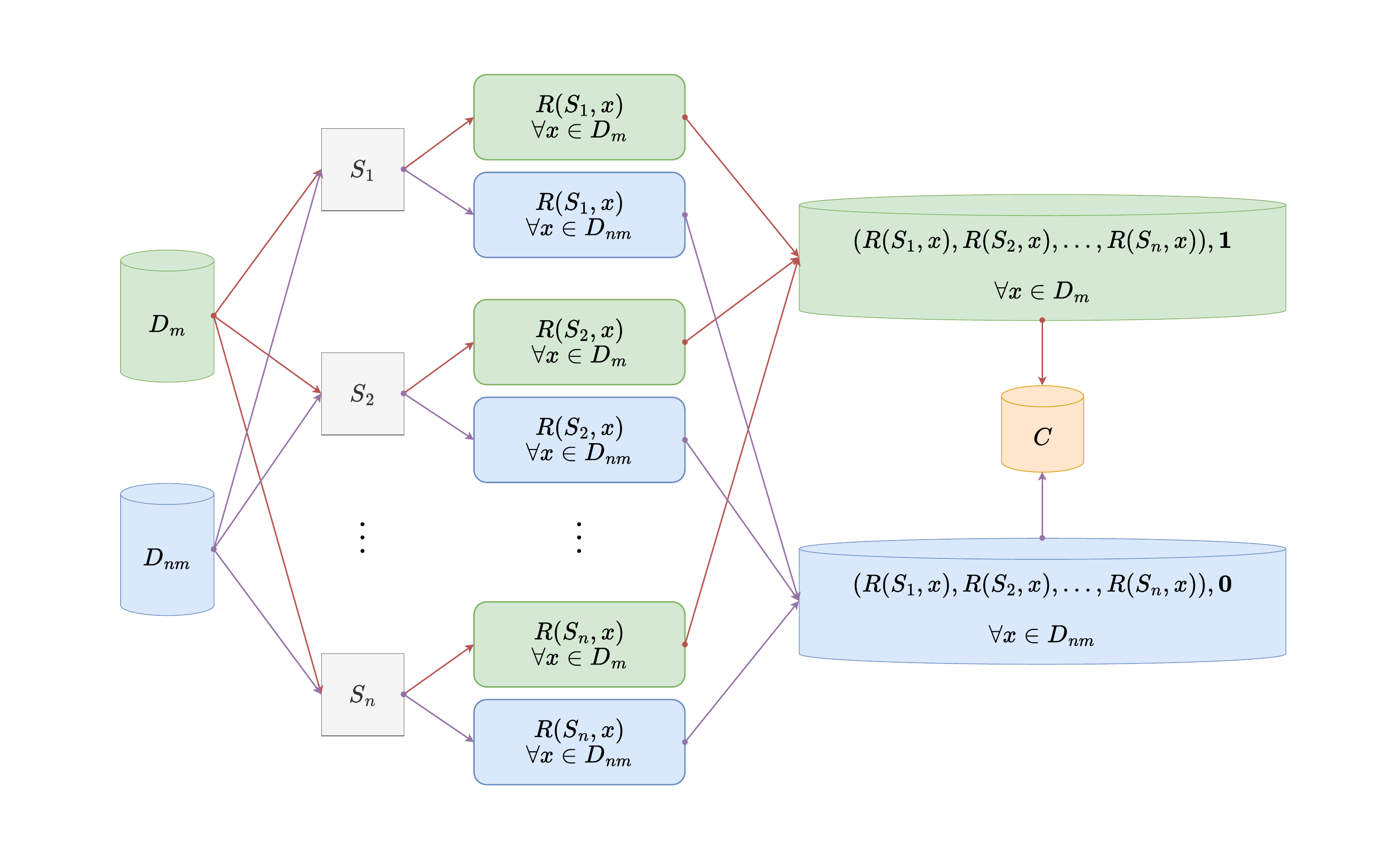}
  \caption{Process for constructing the attack model training dataset.}
  \label{fig:attackscheme}
\end{figure*}

Let $D_m$ and $D_{nm}$ be datasets from the same distribution as $D_{\text{target}}$. In each FL round, we simulate the target model's behavior by training locally on $D_m$ after obtaining a copy of the global model. After $n$ rounds, we have shadow models $S_1, \ldots, S_n$. Let $R(M, x)$ denote some value obtained from feeding a model $M$ with an input $x$, which we expect to differ between training members and non-members (e.g. true class prediction probability). Our goal is to compute these values for each shadow model $S_i$ over both members and non-members to construct a dataset $C$ for training a model capable of inferring membership. The procedure for building dataset $C$ is as follows (see Fig.~\ref{fig:attackscheme}):

\begin{itemize}
\item For each $x \in D_m$, add the vector
$$
\left(R(S_1,x),\dots,R(S_n,x) \right)
$$
to $C$ with label \textbf{1}.
\item For each $x \in D_{nm}$, add the same structure with label \textbf{0}.
\end{itemize}

The reason for this labeling scheme is that we want the attack model to \textbf{return a membership probability}, so we label members with \textbf{1} and non-members with \textbf{0}. Once $C$ is ready, we train a binary classifier $M_\text{attack}$ (e.g., a neural network) to distinguish member and non-member vectors.

\paragraph{Inference Phase}

At this stage, we have a trained model $M_{\text{attack}}$ that predicts membership probability. Suppose we want to assess whether a record $x^\dagger$ belongs to the target private set $D_{\text{target}}$. The corresponding vector is:
\[
s = \left(R(M_1,x^\dagger),\dots,R(M_n,x^\dagger)\right).
\]
Note that this time we use the target models. Feeding $s$ into $M_{\text{attack}}$ yields the estimated probability that $x^\dagger$ belongs to $D_{\text{target}}$.

It is important to highlight that this attack procedure is also applicable in centralized settings ($n=1$), although in such cases we lose access to temporal evolution information.

\subsection{Implementation Details Depending on the Adversary}

As  previously mentioned, we assume adversaries that participate in the federation, i.e., either the aggregator or one of the DOs. The specific implementation of the shadow technique described in Section~\ref{sec:shadowFL} varies across these potential adversaries in several aspects:
\newline
\paragraph{Target Model}

The central aggregator has access to the updated models shared by each DO in every FL round, allowing it to conduct the attack against any of them. Following the notation introduced above, the target models $M_1,\dots,M_n$ correspond in this case to the models shared by the DO that the aggregator aims to attack. On the other hand, each DO only receives the global model obtained after aggregating all individual models from the DOs. In this case, the target models $M_1,\dots,M_n$ are the global models received by the adversarial DO in each FL round. This constraint reduces the impact of the attack, as the information gained is related only to whether data belongs to one of the DOs, without the ability to identify the specific owner (except in the trivial case of two DOs).

\paragraph{Shadow Model and Shadow Dataset}

Both the aggregator and the DOs have white-box access to the models shared in each FL round, enabling them to build and train shadow models externally. However, in the case of an adversarial DO, it may leverage its own real model as a shadow model, since this model actively participates in the collaborative training and may better replicate the target model. Regarding the shadow data used for training these shadow models, since the DO uses its own model, $D_m$ corresponds to its actual dataset employed during training. In contrast, the aggregator does not contribute data to the federation and must therefore obtain it externally. In this work, we assume that the aggregator possesses some knowledge of the data distribution and can generate a synthetic dataset with a distribution similar to that of the target DO. This assumption is more relaxed than in related work \cite{nasr2019comprehensive}, \cite{zari2021efficient}, where the aggregator is assumed to have access to actual data from the victim DO.

\section{Implementing Inference Attacks in Federated Learning}
\label{sec:evaluation}

When studying privacy leakage in deep learning models, \cite{nasr2019comprehensive} observed that the gradients of the loss function with respect to model parameters exhibit significantly different behaviors between training and non-training data. As explained in Section \ref{sec:shadowtechnique}, they introduced a MIA that involves computing ---among many other operations--- the gradients with respect to each parameter of every layer in the model. However, this approach results in a computationally expensive attack. Moreover, one of its assumptions is that the adversary has access to part of the private dataset, which is unrealistic in most cases.

In this work, we propose a novel MIA for federated learning settings that aims to reduce the computational requirements of current shadow-training-based methods, while maintaining similar performance to other state-of-the-art attacks. The proposed approach is based on the observation that the final layers of deep learning models tend to store more information about the training data than intermediate layers. Compared to many other reference attacks, including CS-MIA, the attack we propose is applicable to a broader range of scenarios. For instance, in regression problems (unlike in classification) the model output is a single value rather than a probability vector, which makes many existing attacks, such as CS-MIA, inapplicable. However, as long as the regression model is optimized using gradients, our attack remains feasible.

Below, we describe the two types of attacks implemented and evaluated in this work: CS-MIA, and our novel proposal based on observing the gradients of the last layer of a neural network, specifically designed to achieve higher computational efficiency in federated settings.

\subsection{CS-MIA}

The \textit{Confidence Series based Membership Inference Attack} (CS-MIA) was introduced by \cite{Gu2022} in the context of federated learning and relies on differences in prediction confidence between training set members and non-members. As discussed in Section \ref{sec:membershipinference}, machine learning models tend to assign higher confidence to predictions made on training samples than on unseen instances. Since multiple FL rounds yield multiple predictions, an adversary can also exploit the temporal evolution of confidence to carry out a membership inference attack. As shown in Figure \ref{fig:flconfidence}, this evolution differs between members and non-members.

CS-MIA uses the true class prediction probability as a confidence measure, following the shadow scheme presented in Section \ref{sec:shadowFL}, by setting
$$
R(M,x) = p_y(x),
$$
where $y$ is the true class of $x$ and $p_y(x)$ is the $y$-th entry in $M(x).$

The binary classifier $M_{attack}$ follows the neural network described in \cite{Gu2022}, consisting of one hidden layer with 64 ReLU units followed by a softmax output layer.

\subsection{Gradient-Based Membership Inference Attack}

As previously mentioned, \cite{nasr2019comprehensive} observed that the gradients of the loss function differ significantly between training and non-training samples. Their attack involved computing gradients with respect to all parameters across all model layers. However, this is computationally demanding, and their approach assumes access to part of the private data.

\begin{figure}[!htb]
    \centering
    \begin{subfigure}[t]{0.49\linewidth}
        \centering
        \includegraphics[width=\linewidth]{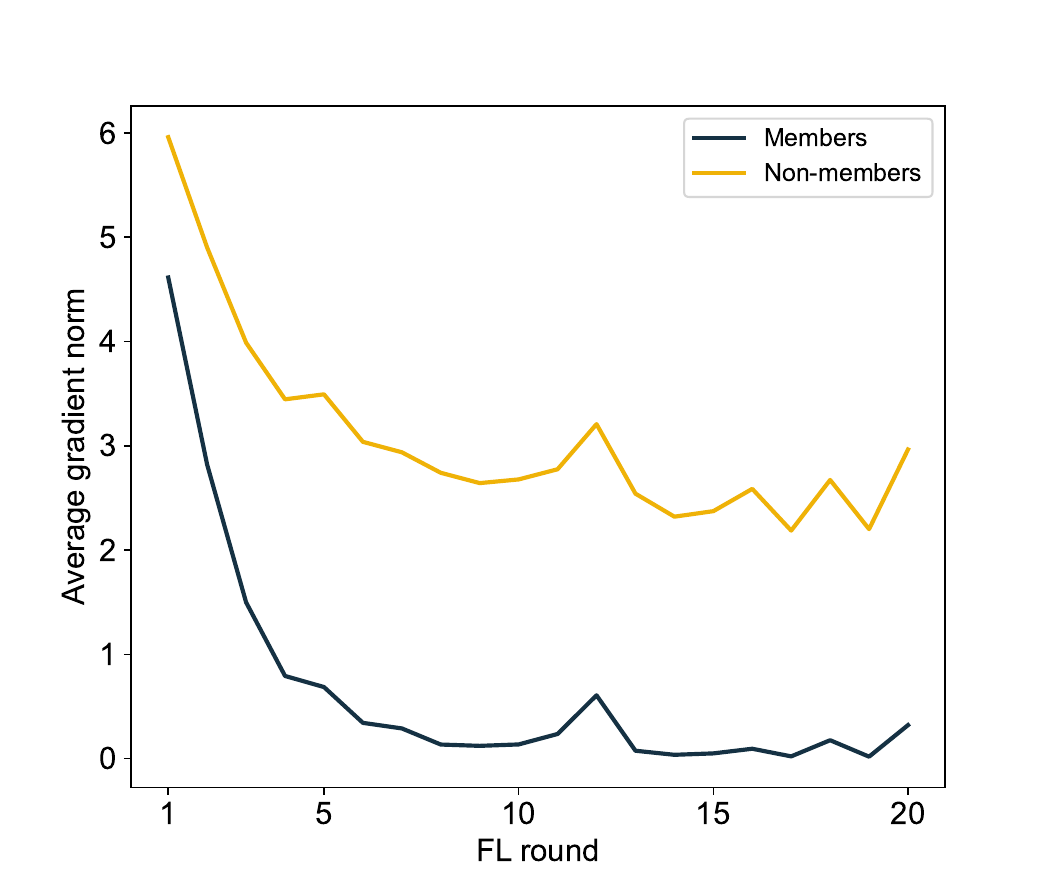}
    \end{subfigure}
    \hfill
    \begin{subfigure}[t]{0.49\linewidth}
        \centering
        \includegraphics[width=\linewidth]{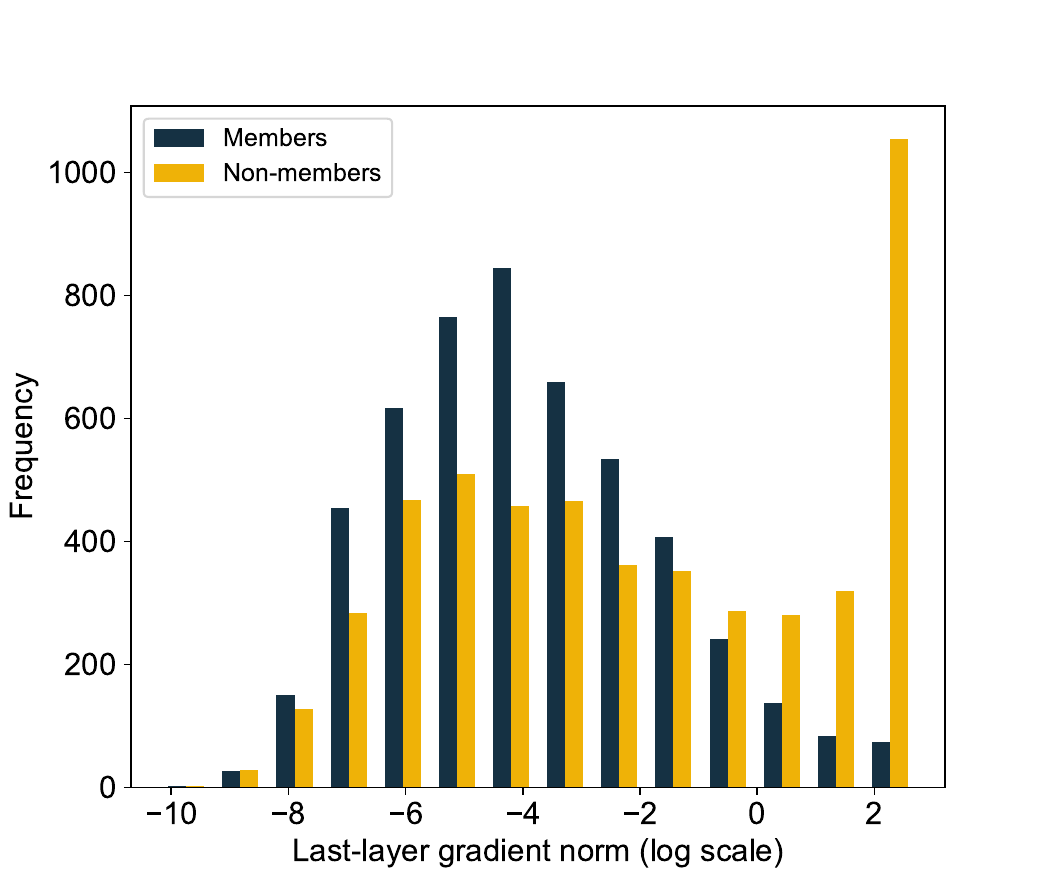}
    \end{subfigure}
    \caption{Evolution of last-layer gradient norms across FL rounds and their distributions in a specific round, comparing members and non-members of the training set. Dataset: Purchase100.}
    \label{fig:gradients-distributions}
\end{figure}

In contrast, we propose a novel attack that reduces computational demands while maintaining effective inference. It is based on the idea that the final model layers store enough training-related information. Figure~\ref{fig:gradients-distributions} illustrates the evolution of last-layer gradient norms across rounds for a specific Data Owner during an FL experiment using the Purchase100 dataset, as well as the distribution of these values in the last FL round. As shown, there are substantial differences in both the distribution and evolution of last-layer gradient norms between members and non-members of the training data. In particular, we can observe how the differences between members and non-members increases at each round of the federated learning process.

We follow the same federated shadow training approach as before, avoiding the need for real private records and requiring only access to a dataset drawn from the same distribution as the training data.

Assume a labeled input $(x, y)$ and let $L(M(x), y)$ be the loss for model $M$ on input $x$ with label $y$. Note that this scenario applies to both classification and regression tasks. We consider the gradient vector $\nabla_{W^*} L(M(x), y)$, where $W^*$ are the parameters of the final layer. The adversary, knowing the model architecture, computes these gradients and their norm:
\[
R(M, x) = \left\| \nabla_{W^*} L(M(x), y) \right\|.
\]

This scalar value serves as input to a binary classifier $M_{attack}$, in this case, a logistic regression model.

This approach drastically reduces computation by focusing only on the last layer's gradients and using logistic regression instead of a neural network. The training set structure of the attack model remains the same as in CS-MIA, since $R(M, \mathbf{x})$ is scalar.

\subsection{Gradient-based Attribute Inference Attack}

Attribute inference attacks, in contrast to MIAs, don't focus on determining the membership of a data point in the training dataset. Instead, they aim to uncover private attributes of data which were part of the training process. For example, an adversary with access to information about certain individuals (such as age, occupation, income...) used to train a model may attempt to infer an unknown private attribute, such as gender.

Based on the technique used in the gradient-based membership inference attack, we derive an attribute inference attack for discrete variables as a natural extension. Since the private attribute the adversary is trying to infer is involved in the training, the model should tend to be more ``surprised'' by incorrect values of the attribute than by correct values. Precisely, suppose we are an adversary with access to a labeled record $(x_1,\dots, x_k;y)$ that was part of the training of a model $M$, but we do not know the value of one of the attributes, for example, $x_k$. For simplicity, suppose the attribute we are trying to infer only takes values $0$ or $1$ and that the actual value of $x_k$ is $0$. Then, over the course of the training rounds, and considering that $M$ will adjust its parameters so that $x_k$ improves the prediction of $y$, the gradients of the loss function with respect to the parameters should be different when evaluated at $(x_1,\dots,0;y)$ and at $(x_1,\dots,1;y)$.

To learn these differences, we will use the same approach already seen for membership inference attacks: incorporating a shadow model into the federation and training it with data for which we do know the value of the private attribute. By doing this, we can build a classifier that allows us to
infer, based on the gradients, what value of the private attribute is correct. It might be noted here that there will be some parameters more influenced by the private attribute than others. However, given the computational cost of analyzing each parameter separately, we will use the same
criterion used in the gradient-based MIA attack: the norm of the gradient with respect to the parameters of the last layer.

Formally, let the training data be of the form $(x,z;y)$, where, for simplicity, we consider the last component $z$ as the private attribute. The scheme described in \ref{sec:shadowFL} for constructing the attack dataset $C$ is adapted as follows: let $D_s$ denote a shadow dataset used to train the shadow models, for which the values of $z$ are known. Let $(R(M,(x,z;y))$ denote the gradient norm as defined for the gradient-based MIA. Then, for each $x\in D_s$, compute
\[
c_k = \left(R(S_1,(x,k;y)),\dots,R(S_n,(x,k;y))\right)
\]
for each possible value $k=0,\dots,K-1$ of the private attribute. Finally, concatenate all $c_k$ vectors to form a feature vector, label it with $z$ (the actual value of the private attribute), and add it to $C$.

By doing this, the adversary gets a new dataset $C$ which will be used to train a classifier $M_\text{attack}$ aiming to distinguish which is the actual value of the private attribute. As the shadow models replicate the behavior of the target models, to infer the value of $z$ for a given record $(x,z;y)$ belonging to the training dataset of the target data owner, the adversary computes the feature vector in the same way as above by replacing $S_i$ with $M_i$ and runs the classifier $M_\text{attack}$ on it to get the predicted value of the private attribute.

\section{Experimental Results and Comparative Analysis}\label{sec:results}

This section presents the evaluation scenario implemented to study the performance of the different inference attacks described in Section~\ref{sec:evaluation}. This analysis has two goals: first, to quantify the effectiveness of each attack using different binary classification metrics (precision, accuracy, recall, F1-score, and AUC-ROC) while also accounting for the computational efficiency of each attack; second, to compare the attacks across different experimental configurations and malicious behaviors from the adversary to draw conclusions about the conditions that promote privacy leakage and the relative robustness of the training mechanisms used.

First of all, we implement three types of adversaries based on their behavior:

\begin{itemize}
    \item \textbf{Agg-SH}. A semi-honest aggregator that captures the updates from the target data owner but does not interfere with the training process.
    \item \textbf{Agg-Mal}. A malicious aggregator that deviates from the protocol by isolating the target data owner, performing a fake aggregation that includes only the shadow model updates and the updates from the target data owner.
    \item \textbf{DO-SH}. A semi-honest data owner that follows the protocol but launches an attack against all other data owners, using its own model as the shadow model and the global model received in each FL round as the target model.
\end{itemize}

Moreover, to evaluate the effectiveness of the implemented attacks and how the dataset used can influence its performance, we consider three different public datasets: for membership inference, we will compare the results obtained in an image classification task (CIFAR-100) with a tabular data classification task (Purchase100). In the case of attribute inference, we employ the Breast Cancer Wisconsin dataset. All models were implemented using PyTorch \cite{Paszke2019} and the evaluation was carried out in a computer with Ubuntu 22.04, Python 3.10, and one NVIDIA H100 GPU with 94Gb of RAM.

\paragraph{CIFAR-100}
This dataset is commonly used for training image classifiers. It was introduced by \cite{KH09} using the \textit{Tiny Images} dataset, which contains 80 million $32\times32$ images. CIFAR-100 comprises 60,000 images in 100 classes, with 600 per class (500 for training, 100 for testing), totaling 50,000 training and 10,000 test images. The model considered in this case is a Convolutional Neural Network based on the architecture proposed by \cite{Salem2018}. It consists of two sequences of \texttt{Conv2D} layers with \texttt{ReLU} activation followed by a 2D pooling layer. In addition, the network includes two fully connected layers with 128 and 100 units, respectively, and a \texttt{tanh} activation function. Regarding the training parameters, we use the Adam optimizer with a learning rate of $10^{-3}$, a weight decay of $10^{-7}$ and a batch size of $128$. The considered FL scenario involves three participants, each holding a local dataset of $10000$ entries. The model is trained locally for five epochs before sharing parameters. The total number of FL rounds is set to 10.

\paragraph{Purchase100}
The Purchase100 dataset was created by \cite{SSSS17} and it consists of customer purchase records represented as binary featured vectors, where each feature indicates whether a specific product was purchased. The dataset includes 600 attributes corresponding to different products, and the records are grouped into 100 classes based on purchasing patterns. In this case, we follow the approach in \cite{nasr2019comprehensive} and use a Fully Connected Network consisting of layers with sizes $600,1024,512,256,128$ and $100$, each followed by a \texttt{ReLU} activation function. We conduct the training using the Adam optimizer with a learning rate of $10^{-3}$ and a batch size of $128$. The FL setting is the same as in the previous case, except that $20$ FL rounds are considered instead of $10$.

\paragraph{Breast Cancer Wisconsin}
To evaluate the attribute inference attack, the Breast Cancer Wisconsin dataset \cite{breast-cancer} was used. It contains 569 samples with 30 features each. These features include measurements such as the radius, texture, perimeter, and area of the cells, among others. The samples are classified into two categories: benign or malignant cancer. Since all variables are continuous, we selected one feature (the area) and, using a K-means algorithm, created two classes. In this way, the area of the tumor serves as a sensitive attribute that the adversary attempts to infer through the described attack. In this case, the federated learning setting has 3 data owners, each contributing 100 data points for training and 50 for testing. The aggregator's shadow model is also trained with 100 entries. The shared ML model is a standard neural network consisting of three fully connected layers of size $30, 16, 6$ and $2$ units, respectively, each followed by a \texttt{ReLU} activation function. For the optimization, Adam optimizer is used with a learning rate of $0.001$, and the model is trained for $5$ local epochs and 10 FL rounds with a batch size of $64$. 

To ensure robustness, each experiment was repeated at least 10 times (depending on the scenario) with different random seeds, introducing partitioning variations and training randomness. For each experiment, we compute the common binary classification metrics: precision, accuracy, recall, F1-Score, and the Area Under the ROC Curves. A classifier with 50\% performance is equivalent to random guessing. Hence, any value above 50\% is interpreted as significant and denotes a successful attack. 

The following subsections detail the results for each scenario and attack, reporting in each case the binary classification metrics, the ROC curves and the computational efficiency. 

\subsection{CS-MIA}

The original results reported by Gu et al.\ \cite{Gu2022} showed outstanding performance for CS-MIA compared to other state-of-the-art attacks. However, after requesting and reviewing their code, we found the following implementation error\footnote{These findings were reported to the authors of CS-MIA in February 2024}. Suppose an FL training process consisting of 3 rounds, so the attacker will have three shadow models, namely $S_1,S_2$ and $S_3$. For simplicity, assume that the shadow dataset only contains two records, and two additional records are taken as non-members. Following the notation presented in Section~\ref{sec:shadowFL}, we may write $D_m=\{x_1, x_2\}$ and $D_{nm}=\{x_3,x_4\}$. In this case, the dataset $C$ used to train the attack model will be as follows, with each row corresponding to one of the records $x_i$:
\[
C=
\left [
\begin{array}{ccc|c}
     R(S_1,x_1) & R(S_2,x_1) & R(S_3,x_1) & 1  \\
     R(S_1,x_2) & R(S_2,x_2) & R(S_3,x_2) & 1  \\
     R(S_1,x_3) & R(S_2,x_3) & R(S_3,x_3) & 0  \\
     R(S_1,x_4) & R(S_2,x_4) & R(S_3,x_4) & 0  \\
\end{array}
\right ]
\]
In the code provided by the paper's authors, this dataset is constructed column-wise. Since each column corresponds to a shadow model, they use a loop that iterates over the models. However, during each iteration, unintentional randomness in sampling when using PyTorch's \texttt{DataLoader} causes the order of inputs to vary. For example, the first iteration might yield the list $(x_2, x_1, x_3, x_4)$, the second iteration $(x_1, x_2, x_4, x_3)$, and the third iteration $(x_2,x_1,x_4,x_3)$. As a result, the dataset $C$ ends up in the form
\[
C=
\left [
\begin{array}{ccc|c}
     R(S_1,x_2) & R(S_2,x_1) & R(S_3,x_2) & 1  \\
     R(S_1,x_1) & R(S_2,x_2) & R(S_3,x_1) & 1  \\
     R(S_1,x_3) & R(S_2,x_4) & R(S_3,x_4) & 0  \\
     R(S_1,x_4) & R(S_2,x_3) & R(S_3,x_3) & 0  \\
\end{array}
\right ]
\]
which is essentially incorrect since each row is mixing confidence scores from different records.

After identifying this error, we re-implemented the attack from scratch and, while the results were positive, they did not reach the levels reported in the original article.

\begin{figure}[!htb]
    \centering
    \begin{subfigure}[t]{0.49\linewidth}
        \centering
        \includegraphics[width=\linewidth]{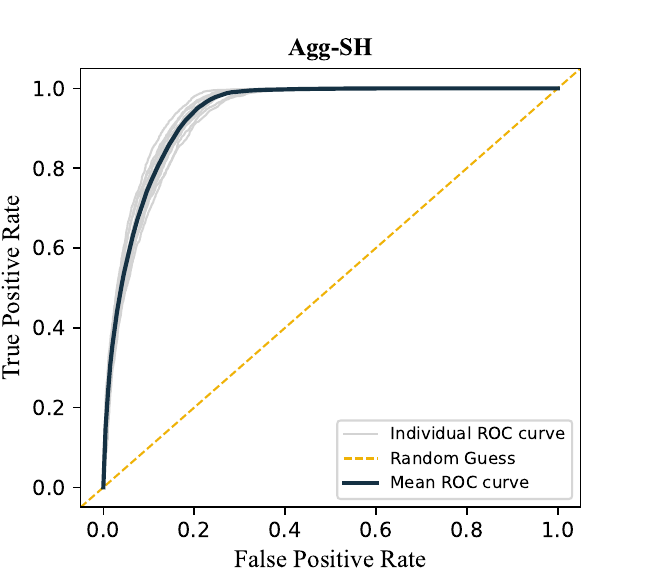}
        \caption{CIFAR-100.}
    \end{subfigure}
    \hfill
    \begin{subfigure}[t]{0.49\linewidth}
        \centering
        \includegraphics[width=\linewidth]{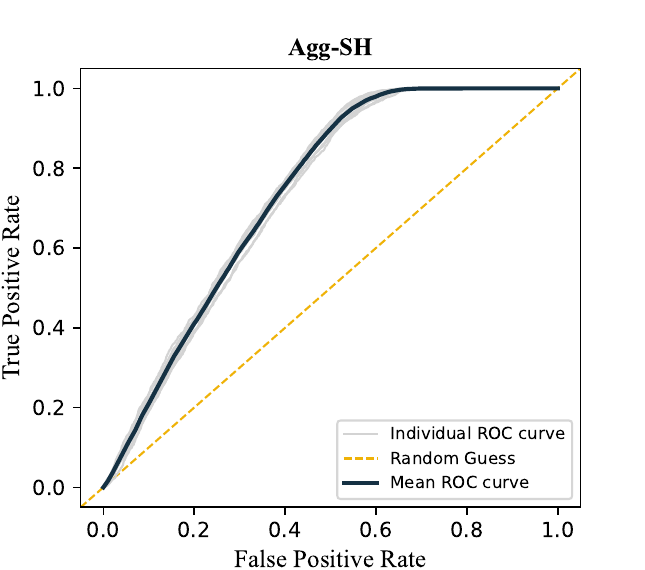}
        \caption{Purchase100.}
    \end{subfigure}
    \caption{ROC curves for the CS-MIA attack for a semi-honest aggregator.}
    \label{fig:roc-csmia-agg-sh}
\end{figure}

\begin{figure}[!htb]
    \centering
    \begin{subfigure}[t]{0.49\linewidth}
        \centering
        \includegraphics[width=\linewidth]{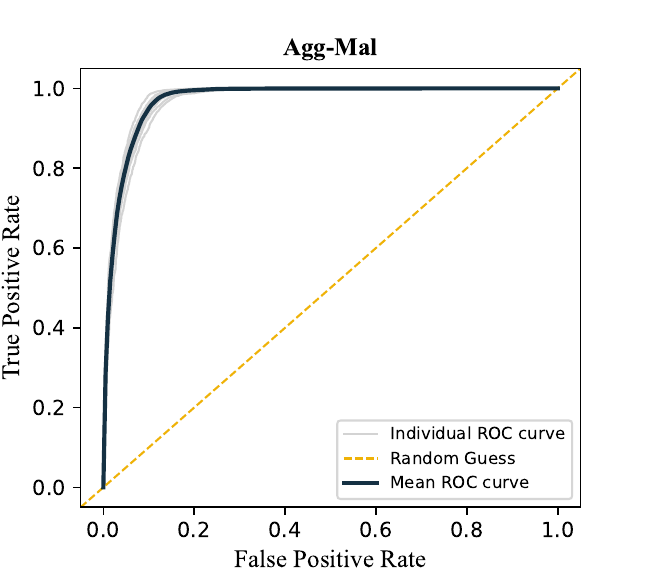}
        \caption{CIFAR-100.}
    \end{subfigure}
    \hfill
    \begin{subfigure}[t]{0.49\linewidth}
        \centering
        \includegraphics[width=\linewidth]{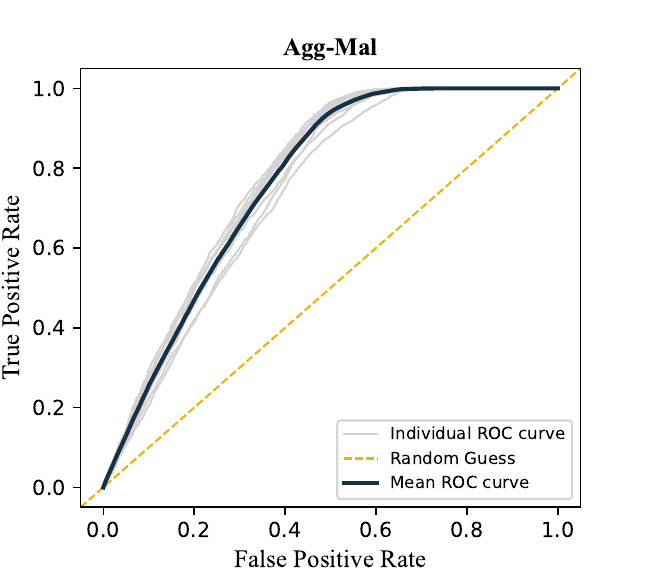}
        \caption{Purchase100.}
    \end{subfigure}
    \caption{ROC curves for the CS-MIA attack for a malicious aggregator.}
    \label{fig:roc-csmia-agg-mal}
\end{figure}

\begin{figure}[!htb]
    \centering
    \begin{subfigure}[t]{0.49\linewidth}
        \centering
        \includegraphics[width=\linewidth]{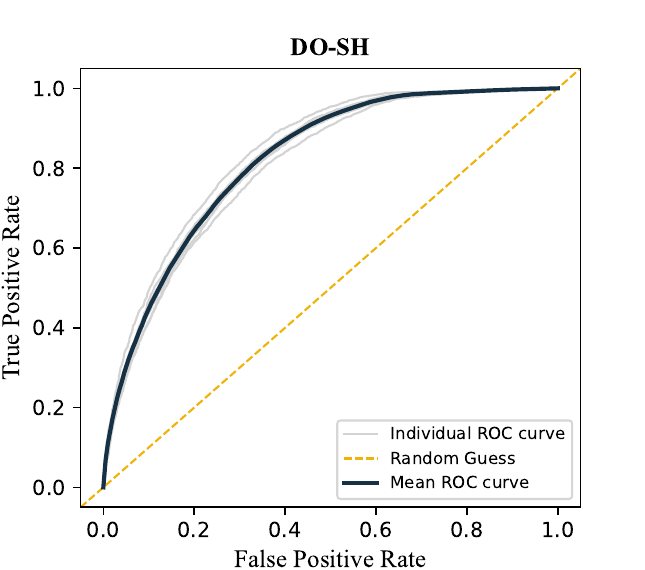}
        \caption{CIFAR-100.}
    \end{subfigure}
    \hfill
    \begin{subfigure}[t]{0.49\linewidth}
        \centering
        \includegraphics[width=\linewidth]{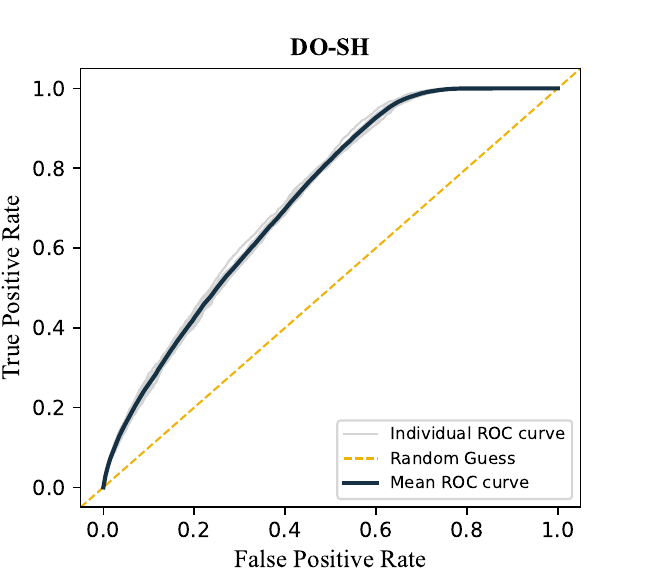}
        \caption{Purchase100.}
    \end{subfigure}
    \caption{ROC curves for the CS-MIA attack for a semi-honest data owner.}
    \label{fig:roc-csmia-do}
\end{figure}

Table~\ref{tab:csmia_cifar100} shows the binary classification metrics (mean and standard deviation over 10 iterations) for each adversary using the CIFAR-100 dataset. Table~\ref{tab:csmia_purchase} displays the results for the Purchase100 dataset. Figs.~\ref{fig:roc-csmia-agg-sh}, ~\ref{fig:roc-csmia-agg-mal} and ~\ref{fig:roc-csmia-do} depict the ROC curves for each experiment iteration and the average ROC curve.

\begin{table}[htbp]
\caption{Results of the CS-MIA attack for each adversary on the CIFAR-100 dataset (mean $\pm$ standard deviation from 10 runs)}
\begin{center}
\begin{tabular}{cccc}
\toprule
Metric & Agg-SH & Agg-Mal & DO-SH\\
\toprule
Accuracy   & $0.87 \pm 0.01$ & $0.93 \pm 0.01$ & $0.74 \pm 0.01$\\
Precision  & $0.83 \pm 0.01$ & $0.89 \pm 0.01$ & $0.71 \pm 0.01$ \\
Recall     & $0.94 \pm 0.02$ & $0.97 \pm 0.01$ & $0.82 \pm 0.02$\\
F1-score   & $0.88 \pm 0.01$ & $0.93 \pm 0.01$ & $0.76 \pm 0.01$\\
AUC-ROC    & $0.93 \pm 0.01$ & $0.97 \pm 0.00$ & $0.82 \pm 0.01$\\
\bottomrule
\end{tabular}
\label{tab:csmia_cifar100}
\end{center}
\end{table}

\begin{table}[htbp]
\caption{Results of the CS-MIA attack for each adversary on the Purchase100 dataset (mean $\pm$ standard deviation from 10 runs)}
\begin{center}
\begin{tabular}{cccc}
\toprule
Metric & Agg-SH & Agg-Mal & DO-SH\\
\toprule
Accuracy   & $0.70 \pm 0.01$ & $0.72 \pm 0.01$ & $0.67 \pm 0.00$\\
Precision  & $0.64 \pm 0.00$ & $0.66 \pm 0.01$ & $0.61 \pm 0.01$ \\
Recall     & $0.93 \pm 0.02$ & $0.92 \pm 0.03$ & $0.90 \pm 0.03$\\
F1-score   & $0.76 \pm 0.01$ & $0.77 \pm 0.01$ & $0.73 \pm 0.01$\\
AUC-ROC    & $0.76 \pm 0.01$ & $0.76 \pm 0.02$ & $0.72 \pm 0.01$\\
\bottomrule
\end{tabular}
\label{tab:csmia_purchase}
\end{center}
\end{table}

Compared with the results originally reported by Gu et al.~\cite{Gu2022}, our re-implementation yields consistently lower metrics. This divergence highlights the importance of reproducibility in privacy research and suggests that subtle implementation details, dataset pre-processing, or adversary assumptions can critically affect attack effectiveness. Importantly, despite this reduction, the attack still achieves strong performance, confirming the underlying vulnerability.

The analysis by adversary type shows clear trends. The malicious Aggregator consistently outperforms the semi-honest version, confirming that protocol deviations strongly amplify privacy risks. Furthermore, the Aggregator adversary is substantially more powerful than the Data Owner, since the latter only observes aggregated updates. These results align with the intuition that access to raw updates from all DOs increases the privacy leakage potential.

At last, dataset dependency also emerges as a relevant factor when studying the performance of a particular membership inference attack. On CIFAR-100, adversary type strongly influences the success of the attack, while on Purchase100, results are similar across scenarios with smaller performance differences. This may be due to differences in data modality: the high-dimensional gradients of deep image neural networks may be able to retain richer membership signals that are captured by the aggregator, whereas gradients from models on tabular datasets reveal a similar amount of membership information which is available even for weaker adversaries with access only to aggregated parameters, such as the data owner.

Finally, the recall values suggest that the attacker can identify most members, although at the cost of lower precision, meaning some false positives are introduced. This imbalance has practical implications, as it may increase the perceived leakage risk in some scenarios.

\begin{table}[htbp]
\caption{Training time (in seconds) and additional time overhead introduced by each CS-MIA attack scenario.}
\begin{center}
\begin{tabular}{ccc}
\toprule
Scenario& Training time & Attack Overhead\\
\toprule
 \multicolumn{3}{c}{CIFAR--100 }\\
\midrule
No attack& 185.00& -\\
Aggregator& 244.00& +32\%\\
DO-SH& 195.55& +5\%\\
\midrule
\multicolumn{3}{c}{Purchase100}\\
\midrule
 No attack& 46.07&\\
 Aggregator& 69.68&+51\%\\
 DO-SH& 55.93&+21\% \\ 
\bottomrule
\end{tabular}
\label{tab:csmia_times}
\end{center}
\end{table}

In terms of computational efficiency Table~\ref{tab:csmia_times} reports the execution times for the scenario without attacks, as well as for the scenarios where either an aggregator or a data owner conducts a CS-MIA. As expected, both attacks increase execution times, but only by a small amount. The difference between the data owner and the aggregator times is due to the fact that the aggregator must train an additional shadow model in each FL round, whereas the data owner uses its own model, which is already participating in the training as shadow model.

\subsection{Gradient-based Membership Inference Attack}

\begin{figure}[!htb]
    \centering
    \begin{subfigure}[t]{0.49\linewidth}
        \centering
        \includegraphics[width=\linewidth]{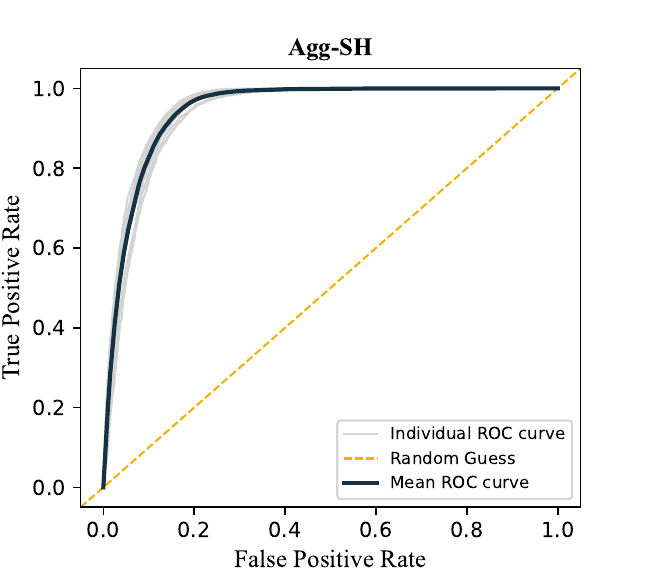}
        \caption{CIFAR-100.}
    \end{subfigure}
    \hfill
    \begin{subfigure}[t]{0.49\linewidth}
        \centering
        \includegraphics[width=\linewidth]{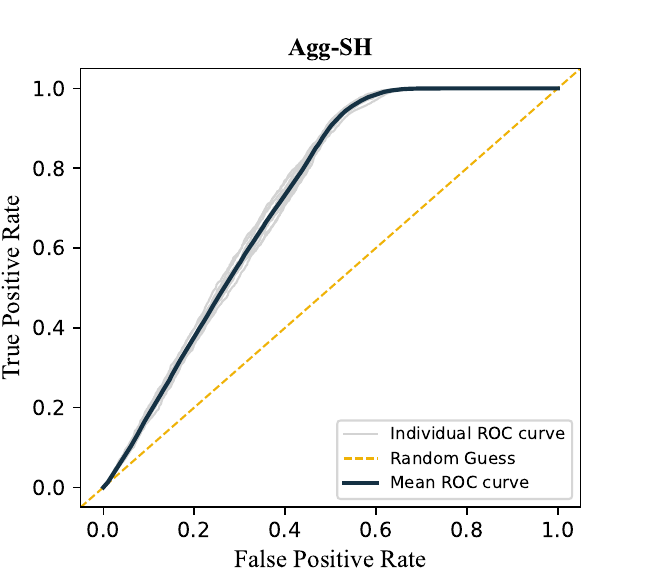}
        \caption{Purchase100.}
    \end{subfigure}
    \caption{ROC curves for the gradient-based attack for a semi-honest aggregator.}
    \label{fig:roc-grad-agg-sh}
\end{figure}

\begin{figure}[!htb]
    \centering
    \begin{subfigure}[t]{0.49\linewidth}
        \centering
        \includegraphics[width=\linewidth]{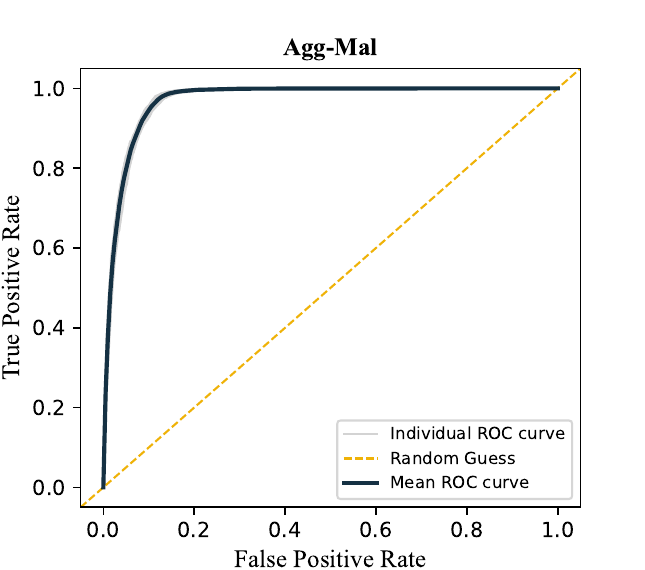}
        \caption{CIFAR-100.}
    \end{subfigure}
    \hfill
    \begin{subfigure}[t]{0.49\linewidth}
        \centering
        \includegraphics[width=\linewidth]{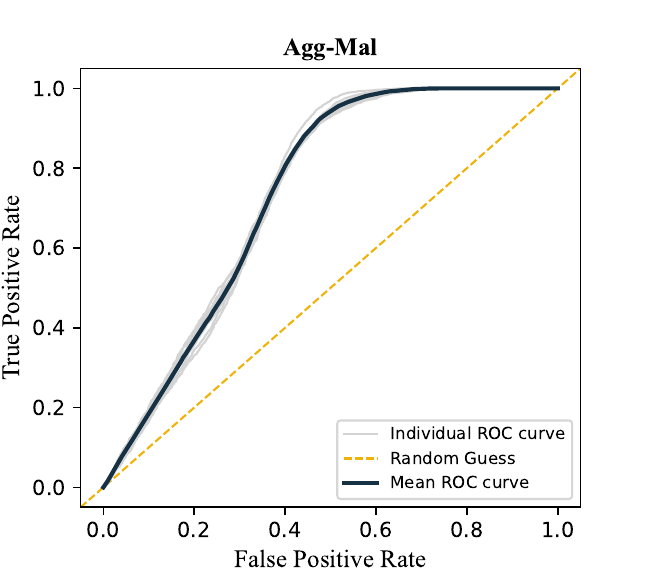}
        \caption{Purchase100.}
    \end{subfigure}
    \caption{ROC curves for the gradient-based attack for a malicious aggregator.}
    \label{fig:roc-grad-agg-mal}
\end{figure}

\begin{figure}[!htb]
    \centering
    \begin{subfigure}[t]{0.49\linewidth}
        \centering
        \includegraphics[width=\linewidth]{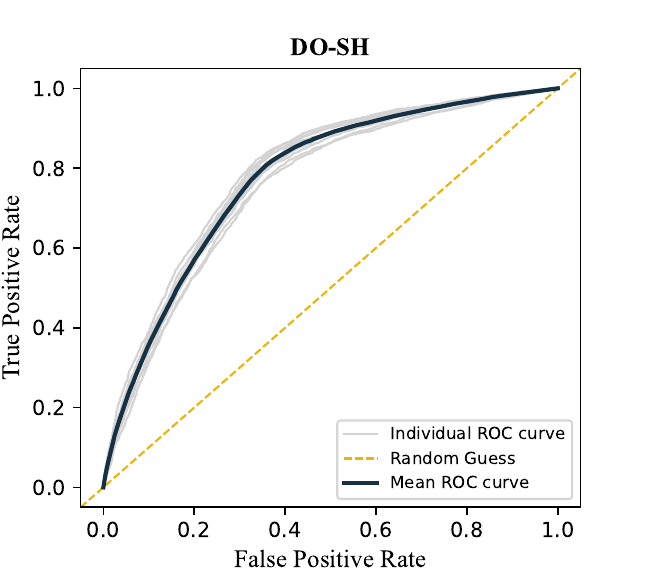}
        \caption{CIFAR-100.}
    \end{subfigure}
    \hfill
    \begin{subfigure}[t]{0.49\linewidth}
        \centering
        \includegraphics[width=\linewidth]{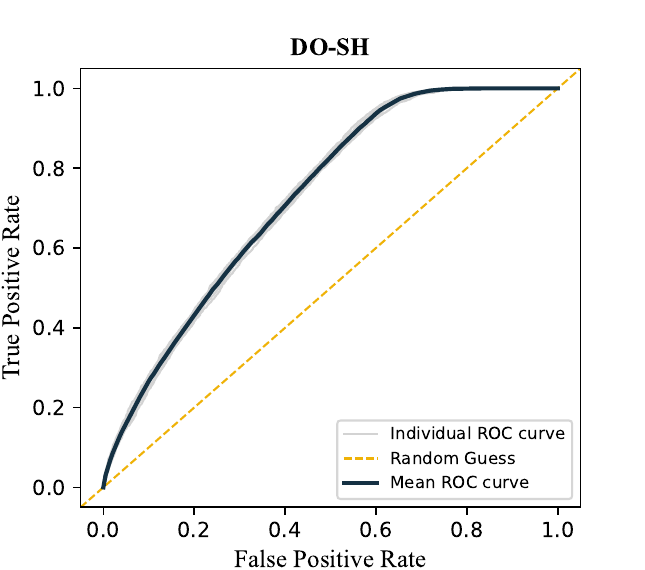}
        \caption{Purchase100.}
    \end{subfigure}
    \caption{ROC curves for the gradient-based attack for a semi-honest data owner.}
    \label{fig:roc-grad-do}
\end{figure}

We evaluate the effectiveness of the gradient-based MIA under the same scenarios as CS-MIA: a semi-honest DO and semi-honest and malicious Aggregator. Tables~\ref{tab:gradient_cifar100} and~\ref{tab:gradient_purchase} show metrics for each dataset, and Figs.~\ref{fig:roc-grad-agg-sh}, ~\ref{fig:roc-grad-agg-mal} and~\ref{fig:roc-grad-do} show the corresponding ROC curves for each adversary.

\begin{table}[htbp]
\caption{Results of the gradient-based attack for each adversary on the CIFAR-100 dataset (mean $\pm$ standard deviation from 10 runs).}
\begin{center}
\begin{tabular}{cccc}
\toprule
Metric & Agg-SH & Agg-Mal & DO-SH\\
\toprule
Accuracy   & $0.89 \pm 0.01$ & $0.93 \pm 0.00$ & $0.73 \pm 0.01$\\
Precision  & $0.85 \pm 0.01$ & $0.89 \pm 0.00$ & $0.70 \pm 0.01$ \\
Recall     & $0.95 \pm 0.01$ & $0.97 \pm 0.00$ & $0.80 \pm 0.02$\\
F1-score   & $0.90 \pm 0.01$ & $0.93 \pm 0.00$ & $0.75 \pm 0.01$\\
AUC-ROC    & $0.95 \pm 0.01$ & $0.97 \pm 0.00$ & $0.78 \pm 0.01$\\
\bottomrule
\end{tabular}
\label{tab:gradient_cifar100}
\end{center}
\end{table}

\begin{table}[htbp]
\caption{Results of the gradient-based attack for each adversary on the Purchase100 dataset (mean $\pm$ standard deviation from 10 runs).}
\begin{center}
\begin{tabular}{cccc}
\toprule
Metric & Agg-SH & Agg-Mal & DO-SH\\
\toprule
Accuracy   & $0.71 \pm 0.00$ & $0.72 \pm 0.01$ & $0.67 \pm 0.00$\\
Precision  & $0.64 \pm 0.00$ & $0.66 \pm 0.00$ & $0.61 \pm 0.00$ \\
Recall     & $0.94 \pm 0.01$ & $0.93 \pm 0.01$ & $0.92 \pm 0.02$\\
F1-score   & $0.76 \pm 0.00$ & $0.77 \pm 0.01$ & $0.74 \pm 0.01$\\
AUC-ROC    & $0.73 \pm 0.01$ & $0.74 \pm 0.01$ & $0.73 \pm 0.01$\\
\bottomrule
\end{tabular}
\label{tab:gradient_purchase}
\end{center}
\end{table}

The gradient-based MIA has a comparable performance to CS-MIA, confirming the robustness of our approach. Interestingly, for semi-honest Aggregators on CIFAR-100, the proposed attack slightly outperforms CS-MIA, which suggests that direct exploitation of gradients provides a more faithful membership signal than the confidence series from CS-MIA. As in CS-MIA, we can observe a similar behavior across datasets, with the gradient-based attack performing also better against deep convolutional neural networks than against smaller, simpler models.

Once again, the Aggregator adversary, and especially the malicious one, achieves the strongest results, highlighting the asymmetric power distribution in classical federated learning scenarios. The ROC curves reinforce these findings, as the area under the curve approaches $0.97$ in the case of CIFAR-100, indicating reliable discrimination capability. Moreover, we can observe lower variability in all the metrics across the 10 executions of the attack when compared to CS-MIA, showcasing the robustness of the proposed attack. 

At last, as in the CS-MIA case, we can observe how the attack performs better against a high-dimensional model (like the deep convolutional neural network implemented with CIFAR-100) than against simpler models trained on tabular data, suggesting that the underlying model and data complexity play a key role in attack performance. 

\begin{table}[htbp]
\caption{Training time (in seconds) and additional time overhead introduced by each gradient-based MIA scenario.}
\begin{center}
\begin{tabular}{ccc}
\toprule
Scenario& Training time & Attack Overhead\\
\midrule 
 \multicolumn{3}{c}{CIFAR--100 }\\
\midrule
No attack& 185.00& -\\
Aggregator& 244.05& +32\%\\
DO-SH& 194.54& +5\%\\
\midrule
\multicolumn{3}{c}{Purchase100}\\
\midrule
 No attack& 46.07&-\\
 Aggregator& 74,65
&+62\%\\
 DO-SH& 60.19&+31\%\\ 
\bottomrule
\end{tabular}
\label{tab:gbmia_times}
\end{center}
\end{table}

Moreover, as in the previous case, we can study the computational efficiency of each attack scenario and dataset. Table~\ref{tab:gbmia_times} reports the execution times for the plain FL without attacks, and the computational overhead introduced by each attack scenario. As in the CS-MIA attack, both attacks increase the execution time of the federated training. It is important to highlight that the computational overhead introduced by the proposed gradient-based MIA is comparable to the CS-MIA attack. Again, as with CS-MIA, the difference between the data owner and the aggregator reported times are due to the fact that the aggregator must train an additional shadow model in each FL round, whereas the data owner uses its own already trained model as shadow model. 

Importantly, the similarity between CS-MIA and the proposed gradient-based attack results shows that our gradient-based design provides comparable effectiveness with a similar computational cost, with the advantage of also offering greater versatility, as it is not tied to classification problems.

\subsection{Gradient-based Attribute Inference Attack}

\begin{figure}[!htb]
    \centering
    \includegraphics[width=0.5\linewidth]{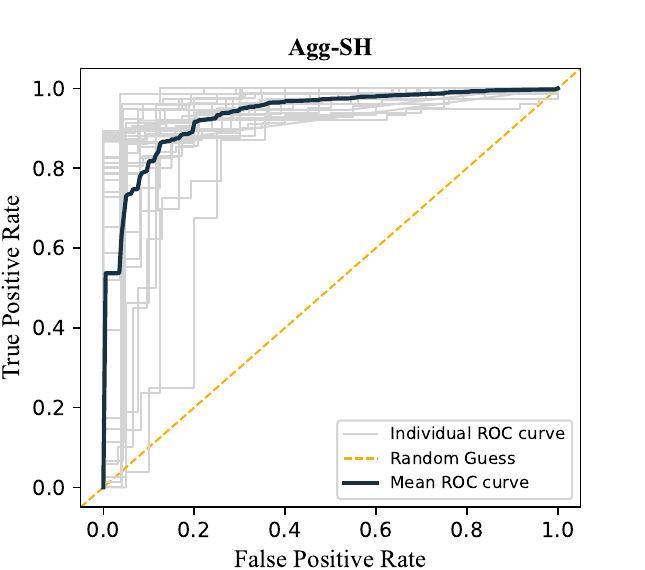}
    \caption{ROC curves for the attribute inference attack.}
    \label{fig:roc-aia}
\end{figure}

In this case, since the sensitive attribute that the adversary tries to infer is binary, we can use the metrics previously employed to assess the effectiveness of the attack. For this scenario, only a semi-honest aggregator is considered as the adversary. Table~\ref{tab:aia} shows the mean and standard deviation after $30$ experiments with different seeds, and Fig.~\ref{fig:roc-aia} shows the ROC curves.

\begin{table}[htbp]
\caption{Results of the attribute inference attack for a semi-honest aggregator on the Breast Cancer dataset (mean $\pm$ standard deviation from 30 runs).}
\begin{center}
\begin{tabular}{cc}
\toprule
Metric & Agg-SH\\
\toprule
Accuracy   & $0.90 \pm 0.03$\\
Precision  & $0.98 \pm 0.03$ \\
Recall     & $0.89 \pm 0.05$ \\
F1-score   & $0.93 \pm 0.03$\\
AUC-ROC    & $0.93 \pm 0.05$\\
\bottomrule
\end{tabular}
\label{tab:aia}
\end{center}
\end{table}

The attribute inference attack reveals particularly severe risks. With accuracy and AUC values of $0.93$, the adversary can infer sensitive features with high reliability. Precision close to $0.98$ is especially alarming, as it implies that when the adversary predicts a positive value of the attribute, it is almost always correct. 

These results are obtained under the semi-honest Aggregator assumption, showing that even moderate adversarial capabilities are sufficient to expose the value of sensitive attributes. This is particularly concerning when applied to medical datasets such as Breast Cancer, where the sensitive attribute is directly tied to an individual’s health status. The stability across $30$ runs further reinforces the robustness of the proposed attack, suggesting that the observed effect is not an artifact of random initialization or sampling.

\begin{table}[htbp]
\caption{Training time (in seconds) and additional time overhead introduced by the attribute inference attack.}
\begin{center}
\begin{tabular}{ccc}
\toprule
Scenario& Training time & Attack Overhead\\
\toprule
No attack& 1.61& -\\
Aggregator& 2.51&56\% \\
\bottomrule
\end{tabular}
\label{tab:gbaia_times}
\end{center}
\end{table}

In terms of computational efficiency, Table~\ref{tab:gbaia_times} presents the execution times for the attribute inference attack compared to the non-attack setting, showing an additional time overhead similar to the MIA cases. This increase is again due to the fact that the aggregator must train a shadow model in each FL round.


\section{Conclusions and Future Work}\label{sec:conclusions}

This paper has implemented and evaluated different inference attacks in federated learning settings, considering both malicious and semi-honest adversaries. We developed statistically grounded attacks---an implementation of CS-MIA and a novel gradient-based inference attack---and demonstrated that, even in federated settings, deep learning models can leak sensitive information if appropriate defenses are not applied. Our approach offers two key advantages: (i) computational efficiency, achieved by targeting final-layer gradient data and using a lightweight classification model to launch the attack; and (ii) generality, since it applies
to any model trained via gradient-based optimization, covering classification, regression, and multi-task settings rather than being limited to classification models.

Considering the adversary type, the Aggregator---with access to the individual updates generated from each private dataset of the DOs---is expected to be more capable of performing successful attacks than a semi-honest DO, who only observes aggregated updates influenced by other participants’ data. Our results for CIFAR-100 confirm this intuition, as both CS-MIA and the gradient-based attack achieve substantially higher performance when the adversary is the Aggregator. This difference becomes even more pronounced when the Aggregator behaves maliciously and deviates from the protocol, resulting in an even more effective attack. Nevertheless, our experiments also show that semi-honest adversaries can still achieve non-trivial success, particularly in the case of attribute inference attacks, which proved highly effective even under weaker assumptions. 

With respect to the comparison of attacks, our results show that the proposed gradient-based membership inference attack achieves very similar performance to CS-MIA, both in terms of accuracy and computational overhead. This confirms that focusing on last-layer gradients is sufficient to capture strong membership signals while maintaining efficiency. Moreover, our approach offers greater versatility: unlike CS-MIA, which is inherently tied to classification tasks and relies on prediction confidence vectors, the gradient-based attack can be applied to any gradient-optimized model, including regression problems, and can be naturally extended to attribute inference.

In terms of computational efficiency, we show how the proposed attack introduces a similar overhead to CS-MIA but remains significantly more efficient than other gradient-based works in the literature, such as the attack proposed in \cite{nasr2019comprehensive}, which requires computing gradients across all layers and relies on partial access to private data. By contrast, our approach drastically reduces computation and memory demands, making it more practical for real-world federated learning deployments.

Finally, our results indicate a relationship between the effectiveness of membership inference attacks and the characteristics of the training dataset. In CIFAR-100, the adversary type plays a significant role on attack performance, with Aggregators achieving a clear advantage over DO. In contrast, on Purchase100, the performance differences between the Aggregator and the DO are smaller. These results could suggest that the high-dimensional gradients obtained from deep convolutional neural networks can preserve more information to carry out the attack, whereas simpler tabular models carry a more uniform amount of information, making them similarly exploitable even by weaker adversaries such as DOs.

Future work will extend this analysis in several directions. First, we aim to study additional inference frameworks, such as reconstruction and other attribute inference attacks, and explore other attacks in FL not based on the shadow technique. Second, we aim to extend our analysis to formally analyze the influence of model and data complexity on the success rate of inference attacks. Finally, we aim to explore the relationship between model interpretability and privacy vulnerabilities, investigating whether interpretable models (e.g., decision trees, linear models, GAMs) are inherently more or less susceptible to membership inference, and whether their transparency can be exploited to design more effective black-box attacks. This line of research seeks to characterize potential trade-offs between interpretability and privacy in machine learning.

\section*{Acknowledgments}
This paper was supported by the TRUMPET project, funded by the European Union under Grant Agreement No. 101070038.



\bibliographystyle{apalike}
\bibliography{biblio}

\end{document}